\documentclass[twoside]{article}

\usepackage[preprint]{aistats2027}
\usepackage{xspace}
\usepackage{booktabs}
\usepackage{graphicx}
\usepackage{multirow}
\usepackage{xcolor}
\usepackage{array}
\usepackage{longtable}
\newcolumntype{L}[1]{>{\raggedright\arraybackslash}p{#1}}
\usepackage{listings}
\lstdefinestyle{prompt}{basicstyle=\ttfamily\scriptsize,breaklines=true,
  columns=fullflexible,keepspaces=true,frame=single,framesep=4pt,
  literate={—}{{---}}1 {…}{{\ldots}}1 {→}{{$\to$}}1 {ℕ}{{$\mathbb{N}$}}1
           {⟨}{{$\langle$}}1 {⟩}{{$\rangle$}}1}

\usepackage[round]{natbib}

\usepackage{amsmath,amsfonts,bm}

\def\eqref#1{equation~\ref{#1}}

\def\1{\bm{1}}

\DeclareMathAlphabet{\mathsfit}{\encodingdefault}{\sfdefault}{m}{sl}
\SetMathAlphabet{\mathsfit}{bold}{\encodingdefault}{\sfdefault}{bx}{n}

\usepackage[hidelinks]{hyperref}
\hypersetup{pdftitle={Definition Grounding in Theory-Level Auto-formalization},
  pdfauthor={Leyan Pan, Christian Szegedy, Frederick Rajasekaran, Robi Bhattacharjee, Gregory Constantine, Wenke Lee}}

\renewcommand{\bibname}{References}
\renewcommand{\bibsection}{\subsubsection*{\bibname}}
\newcommand{\benchname}{ReuseBench\xspace}
\newcommand{\benchinstance}{theory\xspace}

\begin{document}

\runningauthor{Pan, Szegedy, Rajasekaran, Bhattacharjee, Constantine, Lee}

\twocolumn[

\aistatstitle{Definition Grounding in Theory-Level Auto-formalization}

\aistatsauthor{ Leyan Pan$^{1,*}$ \And Christian Szegedy$^{2}$ \And Frederick Rajasekaran$^{2,3}$ \AND
                Robi Bhattacharjee$^{2}$ \And Gregory Constantine$^{1,2}$ \And Wenke Lee$^{1}$ }
\vskip 0.15in
\aistatsaddress{ $^{1}$Georgia Institute of Technology \And $^{2}$AletheAI \And $^{3}$Stanford University } ]
{\renewcommand{\thefootnote}{*}\footnotetext{Work partly done during internship at AletheAI.}}

\begin{abstract}
LLM-based auto-formalization systems can now generate formally verified Lean code at unprecedented rates. However, a formalization advances mathematical research only when further mathematical results can be built on it, especially when its definitions and helper lemmas are general enough to be imported and used idiomatically. We propose definition grounding, where definition formalizations are continuously judged and revised, as a method to ensure formalization quality. We find that agents formalizing definitions from textbook text often produce definitions that later statements cannot use, and only 25\%--50\% of their definition sets are compatible with all later statements. However, clarifying the informal definitions can improve this rate up to 77.5\% averaged over all models. Furthermore, stating a small set of basic properties without proving them can improve model judges' capability to detect definitions with compatibility issues, at only about 11\% of the definition formalization cost. These results are verified with \benchname, a benchmark we curated for formalizing new mathematical fields and theories.
\end{abstract}

\section{Introduction}
AI systems now produce new mathematical results faster than human experts can carefully check their proofs~\citep{openai2026catalog}. As a result, formal verification in proof assistants such as Lean~4 is increasingly important to establishing the correctness of these results. Fortunately, modern AI auto-formalization systems are capable of autonomously formalizing frontier mathematical results, hundreds of pages of mathematical textbooks~\citep{gloeckle2026textbook}, and large-scale proofs spanning several areas of mathematics~\citep{openai2026navierstokes,openai2026catalog}. However, these large-scale formalizations are often overly convoluted, unrefactored, and poorly understood, and they are very hard to use in downstream formalization projects. A formalization's value for advancing mathematical research critically depends on how well it allows further mathematical results to be built on it.

This is clearest when a whole theory is being auto-formalized. Formalizing a new mathematical or theoretical field as a library for others to import starts from its core definitions, then adds basic properties, core results, and eventually progresses to research-level results. Each of these layers is built on earlier ones. We call this setting theory-level auto-formalization following \citet{min2026theorylevel}, which we elaborate on in Section~\ref{sec:theory-autoform}.

In this setting, the most crucial aspect of a correct, reusable formalization is the quality of the formalized definitions. A good formalized definition should be faithful to the intended meaning of the informal source that defines it. It should also be suitable for formalizing downstream statements and their proofs, with the help of its helper lemmas. By \textit{downstream} statements we mean those statements that use the definition (either in their statement or proof). A definition that fails the second condition can go unnoticed through several layers of a theory-level auto-formalization task, and fixing it then means refactoring everything already built on it. 

Writing quality definitions in the manner described above is one of the key failure points of state-of-the-art auto-formalization systems. \citet{ilin2026sorries} report in a case study that only 1 of their 62 definitions was formalized in a correct and reusable way, and \citet{rammal2026scale} report that their agents sometimes replaced a complex object, such as a manifold or a scheme, with an oversimplified definition, resulting in incorrectly formalized downstream statements. 

Thus, in order for large-scale autoformalization projects to create high-quality code, it is necessary to be able to continuously evaluate formalized definitions for reusability. Ideally, an auto-formalization system should be able to evaluate the quality of definitions as they are created, such as through an independent model judge. However, if, throughout the formalization, evidence arises that the original statements have faithfulness or reusability issues, the model should go back and improve them. In this paper we investigate the foundational questions underlying such a framework, namely, when current LLM agents generate good vs.\ bad definition formalizations, and how such definition issues can be identified.

Our main contributions are the following. First, we corroborate the findings above and show that LLM definition formalizations given source text often have issues when being used downstream. Further, we show that these issues often occur deep down in the dependency tree. To address this issue, we propose formalizing a small set of basic properties on top of a candidate definition formalization, and  revising or regenerating the definitions if a judge predicts the definitions to be incompatible with the given evidence. We demonstrate that LLM judges can identify and predict incompatibility problems of definitions, and we show that providing the initial downstream formalizations can increase the accuracy of the LLM judge. To measure the above results, we curate ReuseBench that evaluates whether models write high-quality, reusable Lean code, especially reusable definition formalizations.

\subsection{Related Work}

There has been a large body of work that assesses LLMs' ability to prove theorems \citep{zheng2022minif2f, hu2025minictx, azerbayev2023proofnet}, but much less has been said about their ability to write quality, reusable definitions in the context of entire theories.  \citet{zhang2025definitions} formalize real-world definitions from Wikipedia and arXiv into Isabelle/HOL, and MathAtlas~\citep{patel2026mathatlas} scores definitions alongside theorem statements from 103 graduate textbooks. These works evaluate whether each definition is formalized correctly, while \benchname{} instead tests whether later statements can be built on the formalized definitions. Further, the position paper~\citet{min2026theorylevel} argues that the theory-level formalization approach is the correct one, which the present work focuses on.

A number of works study the ability of LLMs to evaluate the quality of Lean code changes. APE-Bench~\citep{xin2025apebench} evaluates edits to Lean repositories by compilation and LLM judges, and MathlibPR~\citep{xie2026mathlibpr} finds that models cannot tell merge-ready Mathlib pull requests from ones that build but were revised or never merged. Our work focuses on evaluating the ability of LLM judges on definition quality.

Software engineering deals with a similar problem to formalization: passing correctness checks or test cases does not mean being usable. SlopCodeBench~\citep{orlanski2026slopcodebench} measures the accumulative concentration of complexity in functions
while agents extend their own code across 196 checkpoints of 36 problems. The code produced by agents is about twice as complexity-concentrated as that of open-source repositories. SWE-CI~\citep{chen2026sweci}, EvoCode-Bench~\citep{shen2026evocode}, and
MaintainBench~\citep{wang2025maintaincoder} measure how well code survives later changes. \citet{patel2026agentcode} let agents build on agent-written or human-written code, and find resolve rates up to 13.1\% lower on agent-written code, a gap hard to explain by traditional maintainability metrics.
\section{Problem Formulation}

\subsection{Terminology and Notations}
We will follow the following notation and conventions in the rest of the paper:
\begin{itemize}
    \item ``Informal'' refers to mathematical statements and definitions stated in natural language such as in papers and textbooks. ``Formal'' refers to verified code of mathematical statements, such as definitions and proofs in a proof assistant language. The language is Lean 4 unless specified otherwise. ``Formalization'' refers to the process of generating formal counterparts of given informal statements, definitions, and proofs. ``Auto-formalization'' refers to the formalization process done by an AI system.
    \item We use $D$ and $S$ to denote a set of informal definitions and statements (i.e., theorems, lemmas and additional definitions) respectively. We add a dot on top to denote formal, so $\dot{D}$ and $\dot{S}$ denote a set of formal definitions and statements respectively.
    \item $\mathbb{B}$ denotes the boolean values $\{\mathit{True},\mathit{False}\}$.
\end{itemize}

\subsection{Theory-level Autoformalization}
\label{sec:theory-autoform}
Consider the setting of formalizing a new mathematical or theoretical field (i.e., a {\it theory}). The goal is to formalize the core concepts and objects of the theory, followed by basic properties, core lemmas, and finally major theorems. The library is continuously updated with new statements, and external formalization efforts import and use its results. We model this as an iterative process of formalizing $D,S_1,\dots,S_n$, where each iteration corresponds to a formalization session adding new statements to the library or using existing ones. We refer to each of $D,S_1,\dots,S_n$ as a {\it layer}. We use the convention that $S_0=D$.

In autoformalization, the iterative formalization process is mostly done by AI systems such as coding agents, which we represent as $\mathcal{A}$. Suppose that $\dot{C}\in \dot{\mathcal{C}}$ represents all the code within a commit of the formalized library, and $\dot{\mathcal{C}}$ represents the set of all such possible library states. Starting from an empty (besides imported ones like Mathlib) library $\dot{C}_0$, each agent run takes in the current library state $\dot{C}_i$ and is asked to formalize $S_i$ on top of it. This generates an updated repo state $\dot{C}_{i+1}$ that contains formalizations of $D=S_0,S_1,\dots,S_i$. Given the non-determinism of AI agents, we model the process as $\dot{C}_{i+1}\sim \mathcal{A}(S_i|\dot{C}_i)$, where $\mathcal{A}: \mathcal{S}\times\dot{\mathcal{C}}\rightarrow \Delta(\dot{\mathcal{C}})$ is a probability distribution over the next library state, and $\mathcal{S}$ is the set of possible layers.

\paragraph{Refactoring.} One prominent problem in theory-level autoformalization is the need for {\it refactoring}. This happens when, at some stage later in the formalization process, it is realized that earlier definitions and results are {\it incompatible} with the new formalization goals. This means that new statements cannot be stated faithfully using the formal definition as-is, or cannot be proven using the formalized lemmas they are supposed to depend on. This typically happens because of subtly unfaithful or overly narrow formalizations of earlier definitions or statements.

Traditionally, an agent would need to create significant wrappers or coercions of existing formalization results, or completely rewrite the definition or theorems that are the root cause of the issue. This would also require modifying all the proofs and/or statements that are affected by the corrections in the dependency chain, like modifying the proof of a theorem that relies on the changed definition. Refactoring significantly affects the cost of the formalization process, and often lowers the quality of the final formalized library due to the introduced wrappers and coercions. Notably, this impact on cost and quality is more severe when the incompatibility is discovered further into the formalization process, since the dependency chain that needs to be modified becomes larger. Our proposed definition grounding instead aims to ensure the quality of the definitions at every iteration to avoid such large-scale refactoring or code degradation.

\subsection{Definition Grounding}
\label{sec:defground}
To ensure a faithful and clean library of a formalized theory, the formalization quality of the core definitions is of utmost importance. Most results within the theory state properties about them, and a narrowly or unfaithfully formalized definition can be incompatible with statements much further down in the formalization process (i.e., a definition formalization $\dot{D}$ is incompatible with $S_i$ for large $i$). As discussed in the previous section, such refactoring cost would be large and significantly affect codebase quality.

Therefore, if we can predict the formalized definitions that will be incompatible with future downstream statements, we can change them early on and avoid these costly refactorings. This prediction can be done by an LLM judge at each iteration $i$. Given the context $\dot{C}_i$, the judge is asked whether $\dot{D}$ will be compatible with future statements. Operationally, this process can be represented as follows. At each iteration $i$:
\begin{itemize}
    \item The agent $\mathcal{A}$ generates $\dot{C}_{i+1}\sim \mathcal{A}(S_i|\dot{C}_i)$.
    \item We ask the LLM judge $J:\dot{\mathcal{C}}\rightarrow \mathbb{B}$ whether $\dot{C}_{i+1}$ will remain compatible with future statements, denoted $J(\dot{C}_{i+1}) \in \mathbb{B}$.
    \item If $J(\dot{C}_{i+1})=\mathit{False}$, refactor or re-generate $\dot{C}_{i+1}$ until $J(\dot{C}_{i+1})=\mathit{True}$.
    \item Proceed to the next iteration $i+1$ with $\dot{C}_{i+1}$ as the current library state.
\end{itemize}

We refer to the process of producing and maintaining definition formalizations that later layers can be built on {\it definition grounding}\footnote{This term is also used by other works such as \citet{zhang2025definitions} for alternative meanings. We do not claim our term definition to be universal.}. The necessity and effectiveness of the above processes hinge on the following questions:
\begin{itemize}
    \item {\bf Q1:} How often does current AI generate definition formalizations that are incompatible with downstream statements?
    \item {\bf Q2:} Can model judges correctly predict which definition formalizations are likely to cause future refactoring?
    \item {\bf Q3:} Do additional downstream formalizations help model judges better predict the quality of definitions?
\end{itemize}

Our experiments in Section~\ref{sec:expts} focus on answering each of the above questions.

\section{Dataset for Theory-level Auto-formalization}
\begin{table*}[t]
\centering
\caption{Number of definitions in $D$ and of statements in each layer $S_1,\dots,S_5$ of every \benchinstance in \benchname.}
\label{tab:theories}
\begin{tabular}{llrrrrrr}
\toprule
Theory & Source & $D$ & $S_1$ & $S_2$ & $S_3$ & $S_4$ & $S_5$ \\
\midrule
Markov chain mixing times & \citet{levin2017markov} & 20 & 24 & 14 & 9 & 14 & 13 \\
Random walks on graphs & \citet{levin2017markov} & 13 & 15 & 12 & 9 & 8 & 5 \\
Curie--Weiss model & \citet{friedli2017statistical} & 11 & 12 & 13 & 8 & 7 & 5 \\
Nets, covering numbers \& singular values & \citet{vershynin2026high} & 9 & 16 & 14 & 6 & 7 & 4 \\
Linear and Reed--Solomon codes & \citet{guruswami2025essential} & 14 & 15 & 13 & 7 & 7 & 6 \\
Symmetric and Schur polynomials & \citet{grinberg2025introduction} & 14 & 16 & 11 & 9 & 8 & 9 \\
Entropy and Ruzsa distance & \citet{tao2026pfr} & 9 & 17 & 12 & 8 & 6 & 6 \\
Price of anarchy & \citet{karlin2017game} & 12 & 11 & 12 & 9 & 8 & 5 \\
\midrule
Total & & 102 & 126 & 101 & 65 & 65 & 53 \\
\bottomrule
\end{tabular}
\end{table*}

To investigate the performance and methods of current AI agents and systems on theory-level autoformalization, one would need a dataset and benchmark that describes the process of formalizing a theory, including formalizing its core objects and definitions. However, most existing formal proving benchmarks focus on measuring the capability of proving theorems with pre-defined formal statements and definitions (if any). Autoformalization benchmarks focus on writing formal versions of informal statements~\citep{poiroux2024reliable,gao2025herald,liu2025rethinking}, but such tasks still contain little or no evaluation on formalizing the definitions underlying them. 

We therefore propose \benchname as a task of autoformalizing new mathematical fields or theories. Each \benchinstance contains a sequence of incremental tasks of informal definitions and statements to formalize, starting from the core objects and concepts and followed by at most $5$ sets of increasingly advanced statements about the theory. Theories are extracted from introductory textbooks on mathematical or theoretical topics, from which we extract foundational definitions and the ensuing propositions, claims, lemmas, theorems, or exercises about these definitions. \benchname currently contains 8 theories and can be easily extended to include other fields of interest. Most definitions and statements of each theory came from the textbooks as cited in Table~\ref{tab:theories}. We show the full citation list for all layers and theories, including ones from papers, in the supplementary material.

Following the notation in Section~\ref{sec:theory-autoform}, we label these layers as $D,S_1,S_2,\dots,S_5$.
The difficulty of the downstream layers $S_1,\dots,S_5$ is selected roughly according to the following tiers, which are designed to resemble the order of results formalized in real-world formalization projects:
    \begin{itemize}
        \item $D$: Foundational definitions in the textbooks.
        \item $S_1$: Basic properties of the definitions.
        \item $S_2$: Textbook exercises and slightly more advanced properties from the textbook.
        \item $S_3$: Intermediate lemmas towards core results of the theory.
        \item $S_4$: Core results of the theory.
        \item $S_5$: Research-level results, usually from papers rather than textbooks.
    \end{itemize}

Each of these layers is stored as lines in a JSONL file including their informal statement, the source document (i.e., textbook or paper) they came from, which statement in the document they correspond to, and the definitions or statements they depend on for stating and proving from the current or prior layers.

\subsection{Selection Criteria and Cleaning}
\label{sec:selection}
\textbf{The target theories contain non-trivial definitions that are adjacent to Mathlib.} \benchname{} aims to evaluate the capability of agents formalizing new theories from only an empty project based on Mathlib. Therefore, we don't want the theory to be already fully formalized in Mathlib, which would allow the agents to directly import the Mathlib definitions and theorems. On the other hand, many advanced theories are stated based on objects of more basic ones that are also not formalized in Mathlib. Including these would confound the evaluation of formalization of the basic theory versus the more advanced target theory.

To ensure that the theories satisfy these properties, we check definitions against a pinned Mathlib release and ensure that at most a small share
may already have an equivalent there. None of the chapter's
central objects or results may have enough Mathlib support that the downstream statements $S_1,\dots, S_n$ could be stated
and proved without the submitted definitions. At the same time, the definitions'
content must be within reach of what Mathlib already provides, so
that an agent does not first have to build a large missing theory beneath
them.

\textbf{Every definition and statement is self-contained.} Textbooks often state definitions and theorems in ways not suitable for direct formalization. Textbook definitions can often omit implicit edge cases, and both textbook and paper statements can rely on unstated assumptions and conventions. 
Interpreting the precise meanings of informal definitions and theorems is an important part of the challenges of autoformalization. 
A textbook definition can bundle auxiliary claims about the definition that should have been separate theorems. It can also contain references to previous statements by textbook-specific indicators, or directly use notations with attached assumptions introduced earlier in the book. We therefore include restated versions of textbook definitions and theorems alongside their original textbook text. In the cleaned versions, each definition introduces only a
mathematical object and makes no claim about it. When a
statement in the book relies on hypotheses or conventions set earlier in the
chapter, we make them explicit. We also calibrate the definitions and statements using later layers to include missing edge cases and hypotheses that the original text omits. In Table~\ref{tab:one-shot-compat}, we show that this cleaned version substantially improves the formalization quality of the definitions.

\subsection{Content and Data Format}

For each theory, every layer $D,S_1,S_2,\dots,S_5$ is a JSONL file, where each line contains the following fields:
\begin{itemize}
    \item The definition or theorem text, cleaned and restated to be self-contained.
    \item A unique id (e.g., \texttt{def-transition-matrix}), so agents can pinpoint which Lean statements are intended to formalize which informal statement.
    \item The ids of the dependencies that each definition or statement depends on. For example, most theorems in $S_1,\dots,S_5$ depend on at least one definition from $D$.
    \item The verbatim text from the source before cleaning for the definitions in $D$.
    \item Source info, including which textbook/paper source document it corresponds to, which page, lines, or label in the book it refers to.
\end{itemize}

\section{Definition Grounding Experiments}
\label{sec:expts}
\begin{table*}[t]
\centering
\caption{Witness rate in $S_1,\dots,S_5$ and full pass rate of the definition formalizations written by each model, given access to the source text (Source) and the cleaned statements (Cleaned), averaged over all 8 theories. Full pass rate means the percentage of definition formalizations with $L(\dot{D})=\mathit{True}$, i.e., that are fully compatible.}
\label{tab:one-shot-compat}
\begin{tabular}{lcccccc}
\toprule
 & \multicolumn{5}{c}{Witness rate (\%)} & Full pass \\
\cmidrule(lr){2-6}
Model & $S_1$ & $S_2$ & $S_3$ & $S_4$ & $S_5$ & rate (\%) \\
\midrule
\multicolumn{7}{l}{\textit{Source}} \\
GPT 6 Astra & 6.3 & 18.1 & 10.8 & 11.7 & 15.2 & 43.8 \\
GPT 5.6 Sol & 5.2 & 18.0 & 14.1 & 16.9 & 19.2 & 31.2 \\
Claude Fable 5.1 & 4.8 & 10.3 & 5.6 & 6.2 & 3.8 & 50.0 \\
Claude Opus 5 & 5.5 & 6.5 & 6.2 & 9.0 & 6.9 & 50.0 \\
Claude Sonnet 5 & 10.1 & 22.3 & 19.6 & 16.1 & 18.8 & 25.0 \\
\cmidrule(lr){1-7}
All models & 6.4 & 15.0 & 11.3 & 12.0 & 12.8 & 40.0 \\
\midrule
\multicolumn{7}{l}{\textit{Cleaned}} \\
GPT 6 Astra & 0.0 & 1.0 & 0.9 & 0.0 & 2.5 & 81.2 \\
GPT 5.6 Sol & 2.0 & 3.4 & 3.6 & 5.8 & 10.0 & 75.0 \\
Claude Fable 5.1 & 0.0 & 0.0 & 0.0 & 0.0 & 0.0 & 100.0 \\
Claude Opus 5 & 0.4 & 2.5 & 0.0 & 0.0 & 0.0 & 87.5 \\
Claude Sonnet 5 & 8.8 & 13.8 & 11.9 & 12.8 & 25.2 & 43.8 \\
\cmidrule(lr){1-7}
All models & 2.2 & 4.1 & 3.3 & 3.7 & 7.5 & 77.5 \\
\bottomrule
\end{tabular}

\end{table*}

 In this work, we focus on the core questions in definition grounding as outlined in Section~\ref{sec:defground}, which serve as the foundation for high-quality theory-level formalization. 
In Section~\ref{sec:perf-defground}, we evaluate current frontier LLMs on generating definitions that remain usable for stating downstream statements of different levels without significant workarounds or refactoring.
In Section~\ref{sec:expt-judge-pred}, we investigate whether an LLM judge can accurately predict whether a set of formalized definitions can be used idiomatically in unseen downstream statements they will need to support the formalizations of. Then, we investigate whether using a set of definitions to formalize a set of basic downstream statements would help the model judges in making more accurate predictions.

\subsection{Experiment Setup}
\label{sec:setup}
\paragraph{Definition and Downstream Formalizations.} For each theory, a definition generator agent writes Lean files $\dot{D}$ formalizing the informal definitions $D$ of the theory as well as helper lemmas accompanying the definitions. We generate multiple definition candidates using the models described in the Runs paragraph.  We then use a {\it canonical agent} $\mathcal{A}$, using GPT 5.6 Sol, to formalize each of the downstream layers $S_1,\dots,S_5$ sequentially. The process follows the description in Section~\ref{sec:theory-autoform}, where the canonical agent run formalizing $S_i$ starts from library state $\dot{C}_i$ containing formalizations of all previous layers $\{\dot{D},\dot{S}_1,\dots,\dot{S}_{i-1}\}$, and produces $\dot{C}_{i+1}$ additionally containing $\dot{S}_i$. We ask the canonical agent to only {\it state} the downstream statements rather than prove them. Note that the quality of definition formalizations mainly affects whether downstream statements can be stated faithfully and idiomatically, and formal proofs are more affected by the formalization quality of the lemmas they rely on.

\paragraph{Compatibility.} We judge the compatibility of the definition formalization $\dot{D}$ with each downstream statement. Specifically, when formalizing $S_i$, an incompatible definition forces the canonical agent $\mathcal{A}$ to write revised versions of definitions, make wrappers around them, or formalize the statement unfaithfully. We ask the canonical agent to declare any such workarounds it made and mechanically measure how the statements $s\in S_i$ are affected. In these cases, $s$ is called a {\it witness} against $\dot{D}$, denoted $W(s|\dot{D})=\mathit{True}$. In formalization projects, this represents where the agent would likely need to refactor the incompatible definition. To ensure the accuracy of the witness verdicts, we hand-audited a selected set of 20 witnesses and confirmed that they correspond to genuine formalization defects or suboptimal formalization decisions from the definition formalization layer.

\paragraph{Measures.} From the witnesses $W(s|\dot{D})=\mathit{True}$, we can derive three quantities given a definition formalization $\dot{D}$:
\begin{itemize}
    \item The compatibility label $L(\dot{D}|S_1,\dots,S_i)\in \mathbb{B}$, which is True when no statement in $S_1,\dots,S_i$ is a witness against $\dot{D}$. Since our dataset contains 5 layers, the compatibility of $\dot{D}$ over the full downstream $L(\dot{D}|S_1,\dots,S_5)\in \mathbb{B}$ is abbreviated as $L(\dot{D})$.
    \item The witness rate of layer $S_i$, which is the fraction of statements in $S_i$ that are witnesses against $\dot{D}$, $\frac{|\{s\in S_i|W(s|\dot{D})\}|}{|S_i|}$
    \item The full pass rate of a set of definition formalizations, which is the fraction of them with $L(\dot{D})=\mathit{True}$, i.e., with no witness in any layer.
\end{itemize}

\paragraph{Runs.} To generate the definition formalization candidates for each theory, we use 5 different frontier models\footnote{As of September 2026}, GPT 6 Astra, GPT 5.6 Sol, Claude Fable 5.1, Claude Opus 5, and Claude Sonnet 5, to generate diverse definition formalization candidates for each theory. We run 2 repeats of the definition generator agent using each of the above models for each theory, thus 10 definition formalizations for each theory.

\subsection{Can Models Generate Compatible Definition Formalizations?}
\label{sec:perf-defground}
In this section we investigate the performance of frontier models with a standard coding agent harness in generating compatible definition formalizations. Since textbook definitions can have minor flaws for clear informal prose (Section~\ref{sec:selection}), whether agents access sufficient context and clarified rewrites can significantly affect their definition quality.
We therefore evaluate agents under different access levels to the informal information about the definitions that they are asked to formalize:

\paragraph{Access to Verbatim Source Statements (source).} Each definition in a textbook can be traced back to its defining text within the source. This can be a standalone definition block similar to those rendered by \texttt{\textbackslash begin\{definition\}} in \LaTeX, or one or more inline sentences describing the characteristics of the definitions in a paragraph. In this evaluation setting, we provide the generator agent with the transcription of the defining text of each definition it is asked to formalize. We also extract and provide the standing conventions introduced earlier in the textbook. However, the definition generator agent would need to infer unconstrained edge cases, unstated assumptions, or minor inconsistencies that are natural parts of real-world textbooks. This represents the more usual real-world textbook formalization setting.

\paragraph{Access to Cleaned Statements (Cleaned).} As mentioned in Section~\ref{sec:selection}, \benchname also contains cleaned versions of definitions and statements. In this setting, agents are given task structures similar to the ``source'' setting but the informal definition statements are replaced with the cleaned versions. 

\paragraph{Metric.} We report the witness rate of each layer and the full pass rate (Section~\ref{sec:setup}) for every frontier model we use as the definition generator. All values are averaged across all 8 theories and 2 repeated definition generator runs. 

\subsubsection{Results}

Table~\ref{tab:one-shot-compat} shows the experiment results in the metrics described in Section~\ref{sec:perf-defground}. We observe the following:

\paragraph{Formalizations of source definitions are often incompatible downstream.} In the ``Source'' setting where the agent is provided with verbatim transcribed book statements, the formalized definitions encounter substantial compatibility issues on the downstream sets. In particular, even the best-performing model we evaluate, Fable 5.1, generates definition formalizations with no compatibility issues only 50\% of the time when given raw textbook statements. This shows the importance of carefully reviewing and filtering definition candidates generated by AI models when formalizing raw source documents.

\paragraph{Definition incompatibility can surface late.} Recall that later-discovered incompatibilities are more costly because one would need to refactor all of the theory formalizations that are already built on top of the formalized definitions. Our results show that the witness rate remains high at later stages $S_3,S_4,S_5$, which indicates that the generated definition candidates can introduce substantial refactoring cost at these later downstream stages.

\paragraph{Cleaned data substantially improves downstream compatibility.} In the ``Cleaned'' setting, the witness rate of incompatible downstream statements decreases substantially compared to the ``Source'' setting, and the generated candidates also pass all 5 downstream sets more often. Therefore, many textbook texts describing the definitions are often not directly suitable for formalization, and it is highly beneficial to carefully clean the textbook statements based on intended downstream use to avoid future refactoring.

\subsection{Can Model Judges Predict Downstream Compatibility?}
\label{sec:expt-judge-pred}

\begin{table*}[t]
\centering
\caption{Accuracy, AUC and Precision of the baseline (all $\mathit{True}$) and of the judges $J_0,\dots,J_4$ against the compatibility label $L(\dot{D})$, over all 8 theories. }
\label{tab:judge-prediction}
\begin{tabular}{lcccccc}
\toprule
 & Baseline & $D$($J_0$) & $+S_1$($J_1$) & $+S_2$($J_2$) & $+S_3$($J_3$) & $+S_4$($J_4$) \\
\midrule
Accuracy & 0.62 & 0.65 & 0.85 & 0.82 & 0.82 & 0.81 \\
AUC & 0.62 & 0.79 & 0.85 & 0.85 & 0.84 & 0.87 \\
Precision & 0.62 & 0.78 & 0.86 & 0.82 & 0.86 & 0.86 \\
\midrule
Formalization cost (USD) & -- & \$4.32 & \$0.27 & \$0.30 & \$0.24 & \$0.26 \\
Judge cost (5 repeats) & -- & \$0.17 & \$0.21 & \$0.27 & \$0.32 & \$0.35 \\
\bottomrule
\end{tabular}

\end{table*}

\begin{figure*}[t]
\centering
\includegraphics[width=\textwidth]{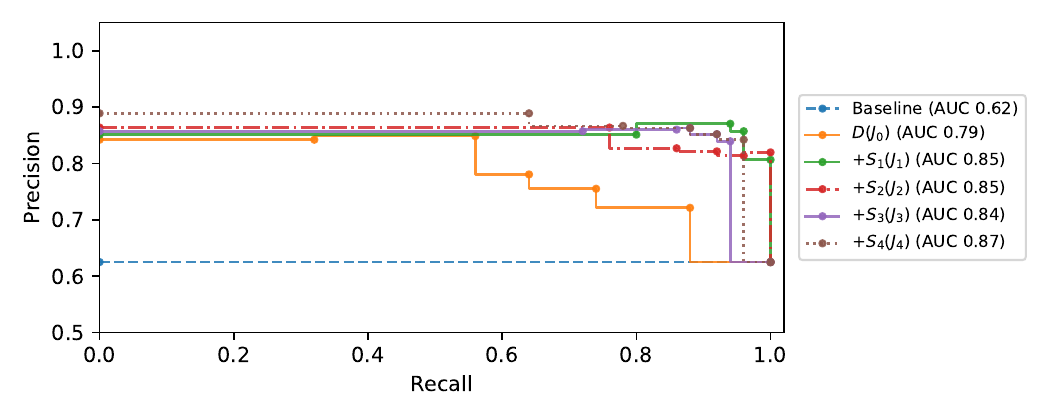}
\caption{Precision-recall curves of the judges $J_0,\dots,J_4$ against the compatibility label $L(\dot{D})$, over all 8 theories, thresholding the number of accepting calls out of 5.}
\label{fig:judge-pr}
\end{figure*}

Since one-shot generation of definition formalizations that are long-term compatible is challenging, a viable strategy is to occasionally review definition quality with an LLM judge based on current partial progress, and revise/refactor the definition formalizations if there are issues (Section~\ref{sec:defground}). However, this is only effective if {\bf Q2:} LLMs can make accurate predictions on which definition formalizations are more compatible with downstream statements. Furthermore, a periodic revision after every layer (as opposed to judging and revising once immediately after formalizing the definitions) is preferable only if {\bf Q3:} additional formalization of downstream statements helps judge models make better predictions.

To investigate both questions, we construct one-shot model judges (using GPT 5.6 Sol) that predict downstream compatibility of formalized definitions $\dot{D}$. The judge takes in the Lean file content containing $\dot{D}$, as well as the informal definition statements, and is asked to predict ``Can future downstream formalizations be built on top of $\dot{D}$ idiomatically and faithfully?''. We denote this judge $J_0=J(\dot{D})\in \mathbb{B}$.
To investigate the second question of whether additional formalizations help make better judgments, we modify the model judge to take in additional context, including the currently formalized downstream statements $\dot{S}_1,\dots,\dot{S}_i$ and their corresponding informal statements $S_1,\dots,S_i$. The model judge is asked instead whether $\dot{D}$ is compatible with the provided $\dot{S}_1,\dots,\dot{S}_i$ evidence as a proxy for future usability. We denote this judge $J_i(\dot{D})=J(\{\dot{D},\dot{S}_1,\dots, \dot{S}_i\})\in\mathbb{B}$. 

For the definition candidates, we use the ``source'' candidates from Section~\ref{sec:perf-defground} for each theory when possible. However, three hard theories do not have any ``source'' candidate without downstream witness, so we use the ``cleaned'' versions for these. This totals 80 candidates with 50 passing.

\paragraph{Metrics.} We use the same compatibility label $L(\dot{D})\in \mathbb{B}$ from Section~\ref{sec:setup} to evaluate the judges $J_0,\dots,J_4$. To investigate {\bf Q2}, we can compare $J_0$ with $L(\dot{D})$ directly. To investigate {\bf Q3}, we are instead interested in whether $J_i$ with $i>0$ makes better predictions than $J_0$.

Given the non-determinism of LLM judges, we perform $k=5$ repeated calls of $J_i(\dot{D})$ for each $i\in\{0,\dots,4\}$. We report three metrics, Accuracy (Acc.), Area Under Precision-Recall Curve (AUC), and Precision. Accuracy straightforwardly measures the probability that the judge aligns with the reference $P(J_i(\dot{D})=L(\dot{D}))$.  

Recall that our goal is to find out and filter/refactor downstream-incompatible definitions as early as possible. Precision measures, if we filter based on whether a majority of the repeated judge calls accepts $\dot{D}$, the probability that a kept $\dot{D}$ is compatible with the downstream sets (i.e., $L(\dot{D})=\mathit{True}$). For any given judge with non-determinism, one can apply the judge's verdicts at different levels of strictness by requiring at least $t$ positive verdicts among $k$ repeated runs. In these cases, we are interested in how well the number of passes among the repeated judge runs differentiates compatible definition formalizations from incompatible ones. Concretely, we calculate the precision-recall curve using the different numbers of passes among repeated judge calls as thresholds, compared with the compatibility label $L(\dot{D})$. We then measure the area under this curve (AUC) as a measure of how differentiating the judge is at any given threshold.

\subsubsection{Results}

\paragraph{Model judges can help eliminate downstream incompatible candidates.} We show that all model judges improve upon the baseline (all $\mathit{True}$) in terms of AUC and precision (Table~\ref{tab:judge-prediction}, Figure~\ref{fig:judge-pr}). Keeping only the definition formalizations that a majority of the definition-only judge's calls accepts raises the downstream compatible rate from $62.5\%$ to $78\%$. For example, filtering with $J_0$ using a majority of 5 votes removes 21 of the 30 incompatible formalizations while discarding 18 of the 50 compatible ones.

\noindent {\bf Early downstream formalizations help judges make better predictions.}
The accuracy, precision, and AUC of the judge improve substantially when we add the downstream formalization evidence. Notably, adding formalized $S_1$ alone raises prediction accuracy from $0.65$ to $0.85$ and AUC from $0.79$ to $0.85$. Adding further downstream evidence $S_2$ to $S_4$ does not substantially improve this figure further. Since $S_1$ only contains basic properties, the average cost for $J_1$, including formalizing $S_1$ and the 5 judge calls, is only $11\%$ of the cost required to formalize the definitions and helpers. Therefore, our experiments suggest that formalizing a small downstream set including basic properties of the definitions and running a model judge to evaluate the quality substantially helps filter out definitions with compatibility issues at a low cost.

Together, the above result supports the definition grounding process of Section~\ref{sec:defground}. Incompatible definition formalizations are common with source-based definition formalization, and a judge given the formalized $S_1$ flags 73\% incompatible definition formalizations in our evaluation before later layers are built on them, at about 11\% of the cost of formalizing the definitions. 

\label{sec:model-judge}

\newpage

\bibliography{iclr2027_conference}

\clearpage
\appendix
\thispagestyle{empty}

\onecolumn
\aistatstitle{Definition Grounding in Theory-Level Auto-formalization: \\
Supplementary Materials}

\section{Sources and Licenses}
\label{app:sources}

{\small
\begin{longtable}{@{}L{2.3cm}L{3.9cm}L{1.9cm}L{3.3cm}L{4.0cm}@{}}
\caption{Detailed source and license of each theory we collected}
\label{tab:sources}\\
\toprule
Theory & Source & Layers & Copy used & License \\
\midrule
\endfirsthead
\toprule
Theory & Source & Layers & Copy used & License \\
\midrule
\endhead
\bottomrule
\endlastfoot
Markov chain mixing times & \citet{levin2017markov} & $D$, $S_1$--$S_5$ & Authors' PDF (2nd ed.) & In copyright; free PDF from the authors \\
 & \citet{diaconis1981generating} & $S_4$ & -- & \textcopyright{} Springer-Verlag \\
 & \citet{bayer1992trailing} & $S_4$ & -- & \textcopyright{} IMS \\
 & \citet{cryan2021modified} & $S_5$ & arXiv:1903.06081v4 & arXiv non-exclusive license \\
 & \citet{ding2010total} & $S_5$ & arXiv:0801.2625v4 & arXiv non-exclusive license \\
 & \citet{basu2017characterization} & $S_5$ & arXiv:1409.3250v4 & arXiv non-exclusive license \\
 & \citet{teyssier2020limit} & $S_5$ & arXiv:1905.08514v1 & arXiv non-exclusive license \\
 & \citet{chen2021optimal} & $S_5$ & arXiv:2011.02075v4 & arXiv non-exclusive license \\
\midrule
Random walks on graphs & \citet{levin2017markov} & $D$, $S_1$--$S_4$ & Authors' PDF (2nd ed.) & In copyright; free PDF from the authors \\
 & \citet{lyons2016probability} & $S_3$, $S_4$ & Authors' PDF, 19 Aug 2026 & \textcopyright{} the authors; free PDF from the authors \\
 & \citet{spielman2011graph} & $S_5$ & arXiv:0803.0929v4 & CC BY 3.0 \\
 & \citet{ding2012cover} & $S_5$ & arXiv:1004.4371v5 & arXiv non-exclusive license \\
 & \citet{dembo2004cover} & $S_5$ & arXiv:math/0107191v2 & None stated on arXiv \\
 & \citet{zhai2018exponential} & $S_5$ & arXiv:1407.7617v1 & arXiv non-exclusive license \\
\midrule
Curie--Weiss model & \citet{friedli2017statistical} & $D$, $S_1$--$S_4$ & Authors' PDF, 22 Aug 2017 & \textcopyright{} the authors; free PDF from the authors \\
 & \citet{chatterjee2007stein} & $S_4$ & arXiv:math/0604352v2 & None stated on arXiv \\
 & \citet{chatterjee2011nonnormal} & $S_5$ & arXiv:0907.4450v3 & arXiv non-exclusive license \\
 & \citet{eichelsbacher2010stein} & $S_4$, $S_5$ & arXiv:0908.1909v1 & arXiv non-exclusive license \\
 & \citet{chen2013stein} & $S_5$ & arXiv:0911.5373v4 & arXiv non-exclusive license \\
 & \citet{can2017cramer} & $S_5$ & arXiv:1709.04267v2 & arXiv non-exclusive license \\
\midrule
Nets, covering numbers \& singular values & \citet{vershynin2026high} & $D$, $S_1$--$S_4$ & Author's PDF, 22 Jul 2026 & \textcopyright{} the author \\
 & \citet{vershynin2012introduction} & $S_4$ & arXiv:1011.3027v7 & arXiv non-exclusive license \\
 & \citet{bandeira2016sharp} & $S_4$ & arXiv:1408.6185v4 & arXiv non-exclusive license \\
 & \citet{rudelson2008littlewood} & $S_5$ & arXiv:math/0703503v2 & None stated on arXiv \\
 & \citet{rudelson2009smallest} & $S_5$ & arXiv:0802.3956v4 & arXiv non-exclusive license \\
 & \citet{livshyts2021smallest} & $S_5$ & arXiv:1909.04219v4 & arXiv non-exclusive license \\
 & \citet{artstein2004duality} & $S_5$ & arXiv:math/0407236v1 & None stated on arXiv \\
\midrule
Linear and Reed--Solomon codes & \citet{guruswami2025essential} & $D$, $S_1$--$S_4$ & Draft, 26 Aug 2025 & CC BY-NC-ND 3.0 \\
 & \citet{ball2019arcs} & $S_2$, $S_5$ & arXiv:1908.10772v2 & arXiv non-exclusive license \\
 & \citet{bao2014equivalent} & $S_4$ & arXiv:1401.3582v2 & arXiv non-exclusive license \\
 & \citet{zhu2024bweight} & $S_3$, $S_4$ & arXiv:2107.01717v4 & CC BY-SA 4.0 \\
 & \citet{chakraborty2026harmonic} & $S_3$ & arXiv:2608.15864v2 & arXiv non-exclusive license \\
 & \citet{shangguan2020combinatorial} & $S_2$ & arXiv:1911.01502v1 & arXiv non-exclusive license \\
 & \citet{brakensiek2023generic} & $S_5$ & arXiv:2206.05256v4 & arXiv non-exclusive license \\
 & \citet{alrabiah2025random} & $S_5$ & arXiv:2304.09445v6 & CC BY 4.0 \\
 & \citet{bensasson2020proximity} & $D$, $S_5$ & IACR ePrint 2020/654, 3 Jul 2021 & CC BY \\
\midrule
Symmetric and Schur polynomials & \citet{grinberg2025introduction} & $D$, $S_1$--$S_5$ & arXiv:2506.00738v1 & CC0 1.0 \\
 & \citet{knutson1999honeycomb} & $S_5$ & arXiv:math/9807160v4 & None stated on arXiv \\
 & \citet{knutson2004honeycomb} & $S_5$ & arXiv:math/0107011v2 & None stated on arXiv \\
 & \citet{huh2022logarithmic} & $S_5$ & arXiv:1906.09633v3 & arXiv non-exclusive license \\
 & \citet{lam2007schur} & $S_5$ & arXiv:math/0502446v3 & None stated on arXiv \\
\midrule
Entropy and Ruzsa distance & \citet{tao2026pfr} & $D$, $S_1$--$S_4$ & teorth/pfr, commit 3d78981 & Apache-2.0 \\
 & \citet{tao2010sumset} & $S_4$ & arXiv:0906.4387v5 & arXiv non-exclusive license \\
 & \citet{madiman2010information} & $S_4$ & arXiv:0901.0044v2 & arXiv non-exclusive license \\
 & \citet{gowers2025conjecture} & $S_4$ & arXiv:2311.05762v2 & CC BY 4.0 \\
 & \citet{green2025sumsets} & $S_5$ & arXiv:2306.13403v3 & CC BY 4.0 \\
 & \citet{gowers2026marton} & $S_5$ & arXiv:2404.02244v2 & CC BY 4.0 \\
\midrule
Price of anarchy & \citet{karlin2017game} & $D$, $S_1$--$S_4$ & Authors' PDF, 13 Dec 2016 & Licensed to AMS \\
 & \citet{roughgarden2002bad} & $S_4$ & Author's version, 5 Dec 2001 & \textcopyright{} ACM \\
 & \citet{roughgarden2003price} & $S_2$--$S_4$ & Author's version, 23 Dec 2002 & \textcopyright{} Elsevier \\
 & \citet{roughgarden2015intrinsic} & $D$, $S_2$--$S_5$ & Author's version, 14 Jul 2015 & \textcopyright{} ACM \\
 & \citet{lin2011stronger} & $S_1$--$S_5$ & Published version, author's website & \textcopyright{} SIAM \\
 & \citet{colinibaldeschi2020selfish} & $S_3$, $S_5$ & arXiv:1703.00927v2 & arXiv non-exclusive license \\
\end{longtable}}

\end{document}